# Ensemble $p$-Laplacian Regularization for Remote Sensing Image Recognition

Xueqi Ma[1], Weifeng Liu[1*], *Senior Member, IEEE*, Dapeng Tao[2], and Yicong Zhou[3], *Senior Member, IEEE*

[1]China University of Petroleum (East China), Qingdao, China 266580

*liuwf@upc.edu.cn

[2]Yunnan University, Kunming, China 650091

[3] Faculty of Science and Technology, University of Macau, Macau, China.

Abstract

Recently, manifold regularized semi-supervised learning (MRSSL) received considerable attention because it successfully exploits the geometry of the intrinsic data probability distribution including both labeled and unlabeled samples to leverage the performance of a learning model. As a natural nonlinear generalization of graph Laplacian, $p$-Laplacian has been proved having the rich theoretical foundations to better preserve the local structure. However, it is difficult to determine the fitting graph $p$-Lapalcian *i.e.* the parameter $p$ which is a critical factor for the performance of graph $p$-Laplacian. Therefore, we develop an ensemble $p$-Laplacian regularization (EpLapR) to fully approximate the intrinsic manifold of the data distribution. EpLapR incorporates multiple graphs into a regularization term in order to sufficiently explore the complementation of graph $p$-Laplacian. Specifically, we construct a fused graph by introducing an optimization approach to assign suitable weights on different $p$-value graphs. And then, we conduct semi-supervised learning framework on the fused graph. Extensive experiments on UC-Merced data set demonstrate the effectiveness and efficiency of the proposed method.

1. Introduction

With rapid advances in storage devices and mobile networks, large-scale multimedia data have become available to ordinary users. However, in practical applications, e.g., human action recognition [17] [18] [19], scene classification [20], text categorization [1], and video annotation and retrieval, the labeled samples always insufficient, though vast amounts of unlabeled samples are readily accessible and provide auxiliary information. Semi-supervised learning (SSL) aiming to exploit both labeled data and the structure imposed by the unlabeled data is designed to address such problem. In SSL, it is assumed that nearby samples are likely to share the same label [2] [3] [4] [5] [6]. The manifold regularization [7] is one of the most representative works, which assumes that the geometry of the intrinsic data probability distribution is supported on the low-dimensional manifold.

In order to build better classifiers, the typical MRSSL algorithms exploit the intrinsic geometry of the labeled and unlabeled samples, then naturally captured by a graph. Therefore, researchers pay their attention to build a good graph to capture the essential data structure. Laplacian regularization (LapR) [7] [15] is one prominent manifold regularization based SSL algorithm, which explores the geometry of the probability distribution by using the graph Laplacian. LapR based SSL algorithms have been widely used in many applications. Luo *et al.* [22] employed manifold regularization to smooth the functions along the data manifold for multitask learning. Hu *et al.* [21] introduced graph Laplacian regularization for joint denoising and superresolution of generalized piecewise smooth images. Jiang *et al.* [23] presented a muti-manifold method for recognition by exploring the local geometric structure of samples. Another relatively new prior is the Hessian Regularization (HesR), which has been shown empirically to perform well in a wide range of inverse problems [8] [9] [10]. In comparison to LapR,

Hessian Regularization (HesR) steers the values of function varying linearly in reference to the geodesic distance. In result, HesR can be more accurate to describe the underlying manifold of data. However, Hessian estimation will be inaccurate while it has poor quality of the local fit for each data point [12]. The $p$-Laplacian [13] [14] is nonlinear generalization of general graph Laplacian and has tighter isoperimetric inequality. In particular, Bühler et al. [11] provided a rigorous proof of the approximation of the second eigenvector of $p$-Laplacian to the Cheeger cut which indicates the superiority of graph $p$-Laplacian in local geometry exploiting. In fact, the parameter $p$ of graph $p$-Laplacian is difficult to determine, and it is a critical factor for the performance of graph $p$-Laplacian.

Unfortunately, it cannot define an objective function to choose graph hyperparameters for intrinsic manifold estimation. In general, the cross validation [24] has been widely utilized for parameter selection. But, to the best of our knowledge, this method that selects parameters in a discrete and limited parameter space lacks the ability to approximate the optimal solution tries. Furthermore, the performance of the classification model are weakly relevant to the difference between the intrinsic and approximated manifolds. Thus, the pure cross validation-based parameter selection cannot perform well on model learning. An automatic and fully approximation of the intrinsic manifold of the data distribution will be valuable for the SSL methods.

In this paper, we propose an ensemble $p$-Laplacian regularized method, which combines a series of graph $p$-Laplacian. By a conditionally optimal way, the proposed method learns to assign suitable weights on graphs, and finally construct an optimal fused graph. The fused graph can sufficiently approximate the intrinsic manifold and preserve the local structure of data by the complementation of graph $p$-Laplacian. We build the graph regularized classifiers including support vector machines (SVM) and kernel least squares (KLS) as special cases for remote sensing image recognition. Experiments on the UC-Merced data set [26] validate effect of proposed method compared with the popular algorithms including Laplacian regularization (LapR), Hessian regularization (HLapR) and $p$-Laplacian regularization (pLapR).

The rest of the paper is organized as follows. Section 2 briefly reviews related work on manifold regularization and $p$-Laplacian learning. Section 3 proposes EpLapR. Section 4 presents the EpLapR for kernel least squares and support vector machines. Section 5 provides the experimental results and analysis on UC-Merced data set. Finally, Section 6 gives the conclusions.

2. Related Work

The proposed EpLapR is motivated by MRSSL and $p$-Laplacian learning. This section briefly describes the related works for better understanding.

2.1 Manifold Regularization

Suppose, the labeled samples are $(x, y)$ pairs drawn from a probability distribution $P$, and unlabeled samples $x$ are drawn according to the marginal distribution $P_x$ of $P$. It assumes that the probability distribution of data is supported on a submanifold of the ambient space. In general, the manifold regularization defines a similarity graph over labeled and unlabeled examples and incorporates it as an additional regularization term. Hence, the manifold regularization framework has two regularization terms, one controlling the complexity measures in an appropriately chosen Reproducing Kernel Hilbert Space (RKHS) and the other controlling the additional information about the geometric structure of the marginal. The objective function of the framework is defined as:

$$f^* = arg\ min_{f \epsilon H_K} \frac{1}{l}\sum_{i=1}^{l} V(x_i, y_i, f) + \Upsilon_A \|f\|_K^2 + \Upsilon_I \|f\|_I^2. \tag{1}$$

Where $V$ is a loss function, such as the hinge loss function $max\ [0, 1 - y_i f(x_i)]$ for Support Vector

Machines (SVM). $\| f \|_K^2$ is used to control the complexity of the classification model, while $\|f\|_I^2$ is an appropriate penalty term which approximated by the graph matrix (e.g. graph Laplacian $L$, $L = D - W$, where $W_{ij}$ is weight vector, the diagonal matrix $D$ is given by $D_{ii} = \sum_{j=1}^{n} W_{ij}$) and the function prediction. The parameters $Y_A$ and $Y_I$ control the complexity of the function in the ambient space and the intrinsic geometry, respectively.

2.2 $p$-Laplacian Regularization

As a nonlinear generalization of the standard graph Laplacian, graph $p$-Laplacian has the superiority on local structure preserving. In mathematic community, discrete $p$-Laplacian has been studied in a general regularization framework. In [27], the objective function of a general discrete $p$-Laplacian regularization framework can be computed as follows:

$$f^* = argmin_{f \in \mathcal{H}(V)}\{S_p(f) + \mu\|f - y\|^2\} \quad (2)$$

where $S_p(f) := \frac{1}{2}\sum_{v \in V}\|\nabla_v f\|^p$ is the $p$-Dirichlet form of the function $f$, $\mu$ is a parameter balancing the two competing terms, $y \in \{-1,0,1\}$ depends on labels of vertex $v$.

Bühler and Hein [11] used the graph $p$-Laplacian for spectral clustering and demonstrated the relationship between the second eigenvalue of the graph $p$-Laplacian and the optimal Cheeger cut as follows:

$$RCC \leq RCC^* \leq p(\max_{i \in V} d_i)^{\frac{p-1}{p}} RCC^{\frac{1}{p}} \quad (3)$$

or

$$NCC \leq NCC^* \leq pNCC^{\frac{1}{p}} \quad (4)$$

where $RCC^*$ and $NCC^*$ are the ratio/normalized Cheeger cut values obtained by tresholding the second eigenvector of the unnormalized/normalized $p$-Laplacian, $d_i$ is the degree of vertex $i$, $RCC$ and $NCC$ are the optimal ratio/normalized Cheeger cut values.

In [25], the whole eigenvector analysis of $p$-Laplacian were achieved by an efficient gradient descend optimization approach as:

$$\min_{\mathcal{F}} J_E(\mathcal{F}) = \sum_k \frac{\sum_{ij} w_{ij} |f_i^k - f_j^k|^p}{\| f^k \|_p^p}$$

$$s.t. \ \mathcal{F}^T\mathcal{F} = I. \quad (5)$$

Where $w_{ij}$ is the edge weight, $f^k$ is an eigenvector of $p$-Laplacian, $\mathcal{F} = (f^1, f^2, \cdots, f^n)$ are whole eigenvectors.

Liu *et al.* [28] proposed $p$-Laplacian regularized sparse coding for human activity recognition.

3. Method

The proposed EpLapR approximates the manifold of the data distribution by fusing a set of graph $p$-Laplacian. First, we show the fully approximation of the graph $p$-Laplacian. Then, we propose the EpLapR.

3.1 Approximation of graph $p$-Laplacian

We approximate the graph $p$-Laplacian ($L^p$) by the fully analysis of the eigenvalues and eigenvectors. In [11], the computation of eigenvalue and the corresponding eigenvector on nonlinear operator $\Delta_p^w$ can be solved by the theorem:

The functional $F_p$ has a critical point at $f$ if and only if $f$ is an eigenvector of $\Delta_p^w$, the corresponding eigenvalue $\lambda_p$ is given by $\lambda_p = F_p(f)$. The definition of $F_p$ is given as:

$$F_p(f) = \frac{\sum_{ij} w_{ij}|f_i-f_j|^p}{2\|f\|_p^p} \tag{6}$$

where

$$\|f\|_p^p = \sum_i |f_i|^p.$$

Here $w_{ij}$ is the edge weight. The above theorem serves as the foundational analysis of eigenvectors and eigenvalues. Moreover, we have $F_p(\alpha f) = F_p(f)$ for all real value $\alpha$.

Suppose that the graph $p$-Laplacian has $n$ eigenvectors $(f^{*1}, f^{*2}, \cdots, f^{*n})$ associated with unique eigenvalues $(\lambda_1^*, \lambda_2^*, \cdots, \lambda_n^*)$. According to the above theorem, if we want to get all eigenvectors and eigenvalues of graph $p$-Laplacian, we have to find all critical points of the functional $F_p$. Therefore, we exploit the full eigenvectors space by solving local solution of the following minimization problem:

$$\min_{\mathcal{F}} J(\mathcal{F}) = \sum_k F_p(f^k)$$

$$s.t. \ \sum_i \phi_p(f_i^k)\phi_p(f_i^l) = 0, \ k \neq l \tag{7}$$

where $\mathcal{F} = (f^1, f^2, \cdots, f^n)$.

Rewrite the optimization problem (7), we analyze the full eigenvectors by solving the following graph $p$-Laplacian embedding problem:

$$\min_{\mathcal{F}} J_E(\mathcal{F}) = \sum_k \frac{\sum_{ij} w_{ij}|f_i^k-f_j^k|^p}{\|f^k\|_p^p}$$

$$s.t. \ \mathcal{F}^T\mathcal{F} = I. \tag{8}$$

The gradient of $J_E$ with respect to $f_i^k$ yields the following equation:

$$\frac{\partial J_E}{\partial f_i^k} = \frac{1}{\|f^k\|_p^p}\left[\sum_j w_{ij} \phi_p(f_i^k - f_j^k) - \frac{\phi_p(f_i^k)}{\|f^k\|_p^p}\right]. \tag{9}$$

The problem (8) can be solved with the gradient descend optimization. However, if we simply use the gradient descend approach, the solution $f^k$ might not be orthogonal [25]. So, the gradient is modified in equation (10) in order to enforce the orthogonality.

$$G = \frac{\partial J_E}{\partial \mathcal{F}} - \mathcal{F}\left(\frac{\partial J_E}{\partial \mathcal{F}}\right)^T \mathcal{F} \tag{10}$$

Meanwhile, the full eigenvalue $\lambda = (\lambda_1, \lambda_2, \cdots, \lambda_n)$ can be computed by $\lambda_k = \frac{\sum_{ij} w_{ij}|f_i^k-f_j^k|^p}{\|f^k\|_p^p}$. Finally, the approximate $L^p$ can be computed by $L^p = \mathcal{F}\lambda\mathcal{F}^T$. We summarize the approximation of graph $p$-Laplacian in Algorithm 1. In the algorithm, the step length $\alpha$ is set to be $\alpha = 0.01 \frac{\sum_{ik}|\mathcal{F}_{ik}|}{\sum_{ik}|G_{ik}|}$.

3.2 EpLapR

Consider the MRSSL setting, where two sets of samples X are available, *i.e.*, $l$ labeled samples $\{(x_i, y_i)\}_{i=1}^l$ and $u$ unlabeled samples $\{(x_j)\}_{j=l+1}^{l+u}$, for a total of $n = l + u$ samples. Class labels are given in $Y = \{y_i\}_{i=1}^l$, where $y_i \in \{\pm 1\}$. Typically, $l \ll u$ and we focus on predicting the labels of unseen examples.

According to the manifold regularization framework, the proposed EpLapR can be written as the following optimization problem:

$$f^* = \arg\min_{f \in H_K} \frac{1}{l}\sum_{i=1}^l V(x_i, y_i, f) + \Upsilon_A\|f\|_K^2 + \frac{\Upsilon_I}{n^2}\mathbf{f}^T\mathbf{L}\mathbf{f}. \tag{11}$$

Here, $\mathbf{f}$ is given as $\mathbf{f} = [f(x_1), f(x_2), \cdots, f(x_{l+u})]^T$, $\mathbf{L}$ is the optimal fused graph with $\mathbf{L} = \sum_{k=1}^m \mu_k L_k^p$, s.t. $\sum_{k=1}^m \mu_k = 1$, $\mu_k \geq 0$, for $k = 1, \cdots, m$. Where we define a set of candidate graph p -Laplacian $C = \{L_1^p, \cdots, L_m^p\}$ and denote the convex hull of set A as: $\text{conv } A = \{\psi_1 x_1 + \cdots + \psi_m x_m | \psi_1 + \cdots + \psi_m = 1, x_i \in A, \psi_i \geq 0, i = 1, \cdots, m\}$. Therefore, we have $L \in \text{conv} C$.

To avoid the parameter $\mu_k$ overfitting to one graph [16], we make a relaxation by changing $\mu_k$ to $\mu_k^\gamma$, and obtain the optimization problem as:.

$$f^* = arg\, min_{f \in H_K} \frac{1}{l} \sum_{i=1}^{l} V(x_i, y_i, f) + Y_A \|f\|_K^2 + \frac{Y_I}{n^2} \mathbf{f}^T (\sum_{k=1}^{m} \mu_k^\gamma L_k^p) \mathbf{f}.$$

$$\text{s.t.} \sum_{k=1}^{m} \mu_k = 1, \ \mu_k \geq 0, \ \text{for } k = 1, \cdots, m \qquad (12)$$

Next, we present theoretical analysis for EpLapR.

**Theorem 1:** For $L \in \text{conv} C$, the solution of the problem (12) exists and admits the following representation:

$$f^*(x) = \sum_{i=1}^{l+u} \alpha_i^* K(x_i, x) \qquad (13)$$

which is an expansion in terms of the labeled and unlabeled examples. Where the kernel matrix $K$ with $K_{ij} = K(x_i, x_j)$ is symmetric positive definite.

The representor theorem presents us with the existence and the general form of the solution of (12) under a fixed $\mu$.

So we rewrite the objective function as

$$f^* = arg\, min_{f \in H_K} \frac{1}{l} \sum_{i=1}^{l} V(x_i, y_i, f) + Y_A \|f\|_K^2 + \frac{Y_I}{n^2} \alpha^T K (\sum_{k=1}^{m} \mu_k^\gamma L_k^p) K \alpha.$$

$$\text{s.t.} \sum_{k=1}^{m} \mu_k = 1, \ \mu_k \geq 0, \ \text{for } k = 1, \cdots, m \qquad (14)$$

**Algorithm 1.** The approximate of graph $p$-Laplacian

**Input:** Training examples $X$; Embedding dimension $K$; $p$

**output:** $p$-Laplacian: $L^p$

Step1: Compute data adjacency matrix $W$ and construct graph Laplacian matrix $L$.

Step 2: Decomposition of graph Lapalcian: $L = USU^T$.

**Initialize**: $\mathcal{F} = U(:, 1:K)$

Step 3: **While not converged do:**

$G = \frac{\partial J_E}{\partial \mathcal{F}} - \mathcal{F} \left(\frac{\partial J_E}{\partial \mathcal{F}}\right)^T \mathcal{F}$, where $\frac{\partial J_E}{\partial \mathcal{F}}$ is given by (9)

$\mathcal{F} = \mathcal{F} - \alpha G$

**End**

Step 4: $\lambda_k = \frac{\sum_{ij} w_{ij} |f_i^k - f_j^k|^p}{\|f^k\|_p^p}$

**return:** $L^p = \mathcal{F} \lambda \mathcal{F}^T$

## 4. Example Algorithms

Generally, the proposed EpLapR can be applied to variant MRSSL-based applications with different choices of loss function $V(x_i, y_i, f)$. In this section, we apply EpLapR to KLS and SVM.

### 4.1 EpLapR kernel least squares (EpLapKLS)

By employing the least square loss in optimization problem (14), we can present the EpLapKLS model defined in Equation (15) as follows

$$f^* = arg\, min_{f \in H_K} \frac{1}{l} \sum_{i=1}^{l} (y_i - f(x_i))^2 + Y_A \alpha^T K \alpha + \frac{Y_I}{n^2} \alpha^T K (\sum_{k=1}^{m} \mu_k^\gamma L_k^p) K \alpha.$$

$$\text{s.t.} \sum_{k=1}^{m} \mu_k = 1, \ \mu_k \geq 0, \ \text{for } k = 1, \cdots, m \qquad (15)$$

Then, we obtain the partial derivative of the objective function with respect to $\mu_k$ and $\alpha$ as follows:

$$\frac{\partial f}{\partial \mu_k} = \frac{Y_I}{n^2} \alpha^T \boldsymbol{K}(\gamma \mu_k^{\gamma-1}) L_k^p \boldsymbol{K} \alpha \tag{16}$$

$$\frac{\partial f}{\partial \alpha} = \frac{1}{l}(Y - J\boldsymbol{K}\alpha)^T(-J\boldsymbol{K}) + \left(Y_A \boldsymbol{K} + \frac{Y_I}{n^2} \boldsymbol{K}L\boldsymbol{K}\right)\alpha \tag{17}$$

Here, we adopt a process that iteratively updates $\alpha$ and $\mu_k$ to minimize $f^*$. Firstly, when $\mu_k$ is fixed, (17) turns to $\mathrm{argmin}_{\mu_k} f$, from which we can derive that

$$\alpha^* = (J\boldsymbol{K} + Y_A lI + \frac{Y_I l}{n^2} \boldsymbol{L}\boldsymbol{K})^{-1} Y. \tag{18}$$

Where $I$ is $n*n$ diagonal matrix; $J$ is an $n*n$ diagonal matrix with the labeled points diagonal entries as 1 and the rest 0; $Y$ is an $n$ dimensional label vector given by $Y = \mathrm{diag}(y_1, \ldots, y_l, 0, \cdots, 0)$.

Given fixed $\alpha$, we obtain the solution of $\mu_k$ with s.t. $\sum_{k=1}^{m} \mu_k = 1$:

$$\mu_k = \frac{\left(\frac{n^2}{Y_I \alpha^T \boldsymbol{K} L_k^p \boldsymbol{K} \alpha}\right)^{\frac{1}{\gamma-1}}}{\sum_{k=1}^{m}\left(\frac{n^2}{Y_I \alpha^T \boldsymbol{K} L_k^p \boldsymbol{K} \alpha}\right)^{\frac{1}{\gamma-1}}}. \tag{19}$$

The iterative solution procedure of EpLapR is described in Table 1.

Table 1. Iterative solution method for EpLapR

| |
|---|
| Step1: Initialize $\mu_k \in \boldsymbol{R}^N$. |
| Step2: Update $\alpha$ according to (18). |
| Step3: Based on the updated $\alpha$, re-calculate $\mu_k$ according to (19) |
| Step4: Repeat from Step 2 until convergence. |

Now, we prove the convergence of this iterative solution. Denote by $\alpha^t$ and $\mu^t$ the values of $\alpha$ and $\mu$ in $t^{\mathrm{th}}$ iteration of the process, we have

$$f(\alpha^{t+1}, \mu^{t+1}) \leq f(\alpha^t, \mu^{t+1}) \leq f(\alpha^t, \mu^t).$$

4.2 EpLapR support vector machines (EpLapSVM)

The EpLapSVM solves the optimization problem (14) with the hinge loss function as

$$f(*) = \arg\min_{f \in H_K} \frac{1}{l}\sum_{i=1}^{l}(1 - y_i f(x_i))_+ + Y_A \alpha^T \boldsymbol{K}\alpha + \frac{Y_I}{n^2} \alpha^T \boldsymbol{K}(\sum_{k=1}^{m} \mu_k^\gamma L_k^p)\boldsymbol{K}\alpha. \tag{20}$$

The partial derivative of the objective function with respect to $\mu_k$ is same as equation (16).

Introducing Lagrange multiplier method with $\beta_i$ and $\eta_i$, and add an unregularized bias term $b$, we arrive at a convex differentiable objective function:

$$L(\alpha, \xi, b, \beta, \eta) = \frac{1}{l}\sum_{i=1}^{l}\xi_i + \frac{1}{2}\alpha^T\left(2Y_A \boldsymbol{K} + 2\frac{Y_A}{n^2}\boldsymbol{K}(\sum_{k=1}^{m}\mu_k^\gamma L_k^p)\boldsymbol{K}\right)\alpha$$

$$- \sum_{i=1}^{l}\beta_i\left(y_i\left(\sum_{j=1}^{l+u}\alpha_j \boldsymbol{K}(x_i, x_j) + b\right) - 1 + \xi_i\right) - \sum_{i=1}^{l}\eta_i\xi_i$$

We reduce the Lagrangian using $\frac{\partial L}{\partial b} = 0$ and $\frac{\partial L}{\partial \xi_i} = 0$, and take partial derivative with respect to $\alpha$

$$\alpha^* = (2Y_A I + 2\frac{Y_I}{n^2}\boldsymbol{LK})^{-1} J^T Y \beta^*. \qquad (17)$$

Where $I$ is $l*l$ diagonal matrix; $J$ is an $l*n$ diagonal matrix given by $J = [I, 0]$; $Y$ is an $l$ dimensional label vector given by: $Y = \text{diag}(y_1, \ldots, y_l)$. $\beta^*$ is the $n$-dimensional variable given by

$$\beta^* = \max_{\beta \in \mathbb{R}^l} \sum_{i=1}^{l} \beta_i - \frac{1}{2}\beta^T Q \beta$$

$$\text{subject to:} \sum_{i=1}^{l} \beta_i y_i = 0$$

$$0 \leq \beta_i \leq \frac{1}{l}, \quad i = 1, \cdots, l$$

where

$$Q = YJK(2Y_A I + 2\frac{Y_I}{n^2}\boldsymbol{LK})^{-1} J^T Y.$$

Then, the problem (20) can also be solved by the iterative solution process in Table 1.

5. Experiments

In this section, to evaluate the effectiveness of the proposed EpLapR, we compare EpLapR with other local structure preserving algorithms including LapR, HesR and pLapR. We apply the support vector machines and kernel least squares for remote sensing image classification. Fig. 1 illustrates the framework of EpLapR for UC-Merced data set.

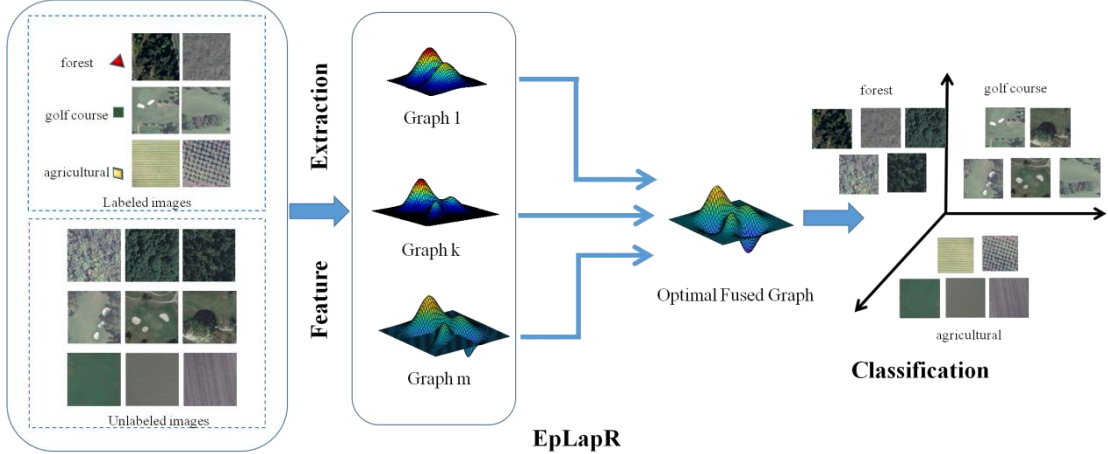

Fig. 1.   The framework of EpLapR for remote sensing image classification.

UC-Merced data set [26] consists of 2100 remote sensing images collected from aerial orthoimage with the pixel resolution of 1 foot. The original images were downloaded from the United States Geological Survey National Map of different U.S. regions. There are totally 21 classes including chaparral, dense residential, medium residential, sparse residential, forest, freeway, agricultural, airplane, baseball diamond, mobile home park, overpass, parking lot, river, runway, beach, buildings, golf course, harbor, intersection, storage tanks, and tennis courts (see in Fig. 2). It is worth noticing that this dataset has some highly overlapped classes, *e.g.*, sparse residential, medium density residential, and dense residential, make it difficult for classification.

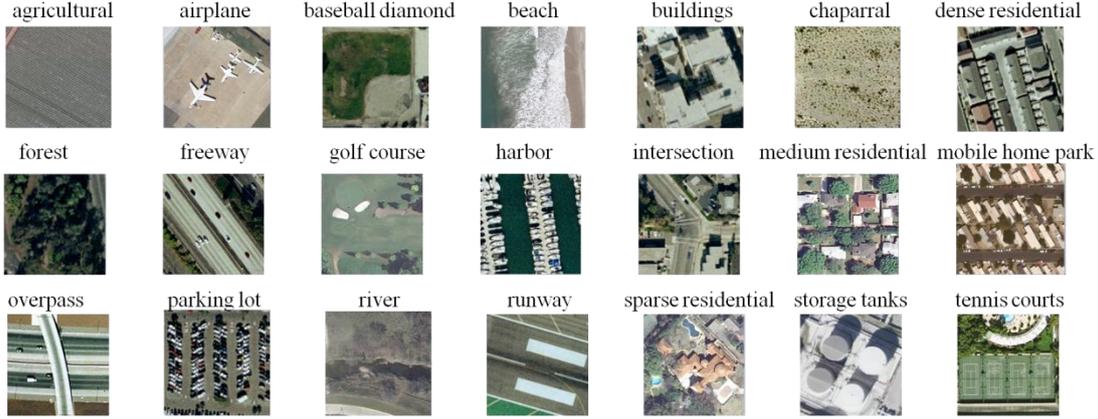

Fig. 2.  Some example images of UC-Merced data set. The dataset totally has 21 remote sensing categories.

In our experiments, we extract high-level visual features using the deep convolution neural network (CNN) [29]. We randomly choose 50 images per class as training samples and the rest as testing samples. In semi-supervised classification experiments, in particular, we select 10%, 20%, 30%, 50% samples of training data as labeled data, and the rest as unlabeled data. To avoid any bias introduced by the random partition of samples, the process is repeated five times independently.

We conduct the experiments on the data set to choose suitable model parameters. The regularization parameters $\gamma_A$ and $\gamma_I$ are selected from the candidate set $\{10^i | i = -10, -9, -8, \cdots, 10\}$ through cross-validation. For pLapR, the parameter $p$ are chosen from $\{1, 1.1, 1.2, \cdots, 3\}$ through cross-validation with 10% labeled samples on the training data. Fig. 3 illustrates the mAP performance of pLapR on the validation set when $p$ varies. The x-axis is the parameter $p$ and the y-axis is mAP for performance measure. We can see that the best mAP performance for pLapR can be obtained when $p = 2.8$. For EpLapR, we created two graph $p$-Laplacian sets. For the first set (EpLapR-3G), we choose $p = \{2.5, 2.7, 2.8\}$, which led to 3 graphs. For another one (EpLapR-5G), with 5 graphs where $p = \{2.4, 2.5, 2.6, 2.7, 2.8\}$. We verify classification performance by average precision (AP) performance for single class and mean average precision (mAP) [30] for overall classes.

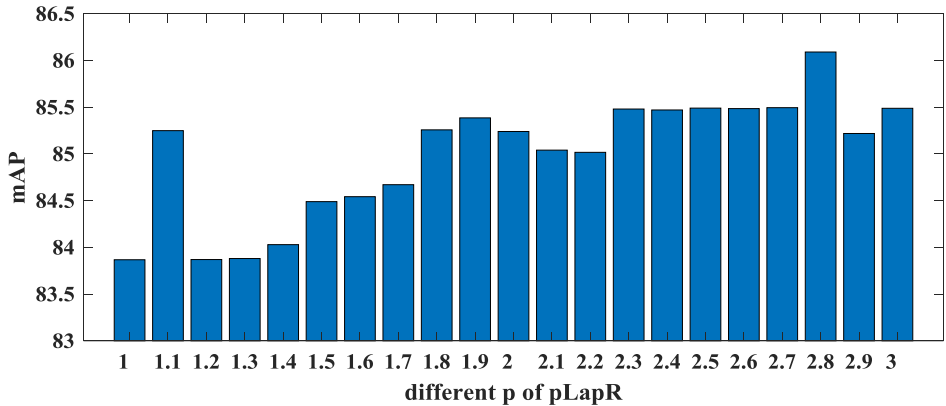

Fig. 3.  Performance of mAP with different $p$ values on validation set.

We compare our proposed EpLapR with the representative LapR, HesR and pLapR. Figure 4 and Figure 5 demonstrate the mAP results of different algorithms on KLS methods and SVM methods, respectively. We can see that, in most cases, the EpLapR outperforms LapR, HesR and pLapR, which shows the advantages of EpLapR in local structure of preserving.

To evaluate the effectiveness of EpLapR for single class, Fig. 6 and Fig. 7 show the AP results of different methods on several selected remote sensing classes including medium residential, parking lot, sparse residential and tennis court. Fig. 6 reveals the KLS method, while Fig. 7 represents the SVM method. We can find that, EpLapR performs better than LapR, HesR and pLapR by sufficiently explore the complementation of graph $p$-Laplacian.

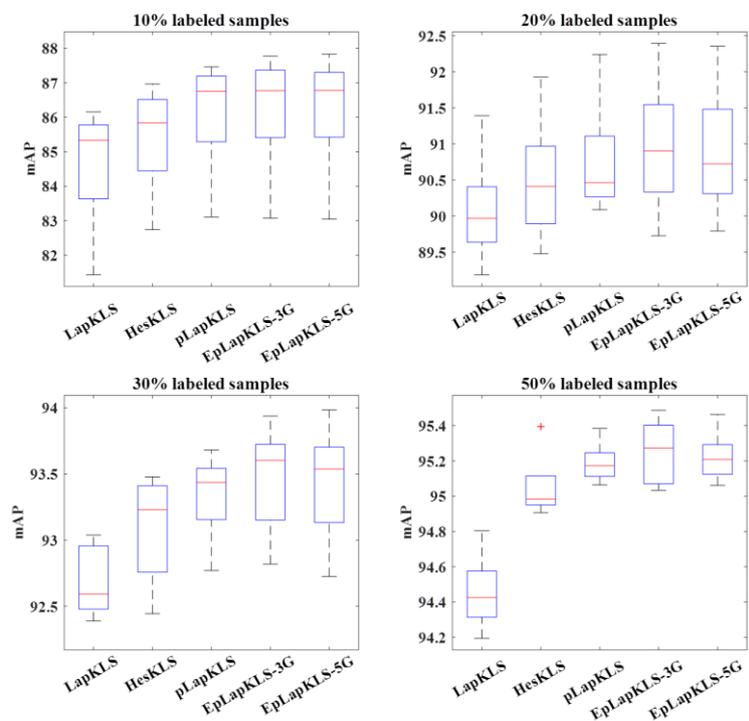

Fig. 4.   mAP performance of different algorithms on KLS method.

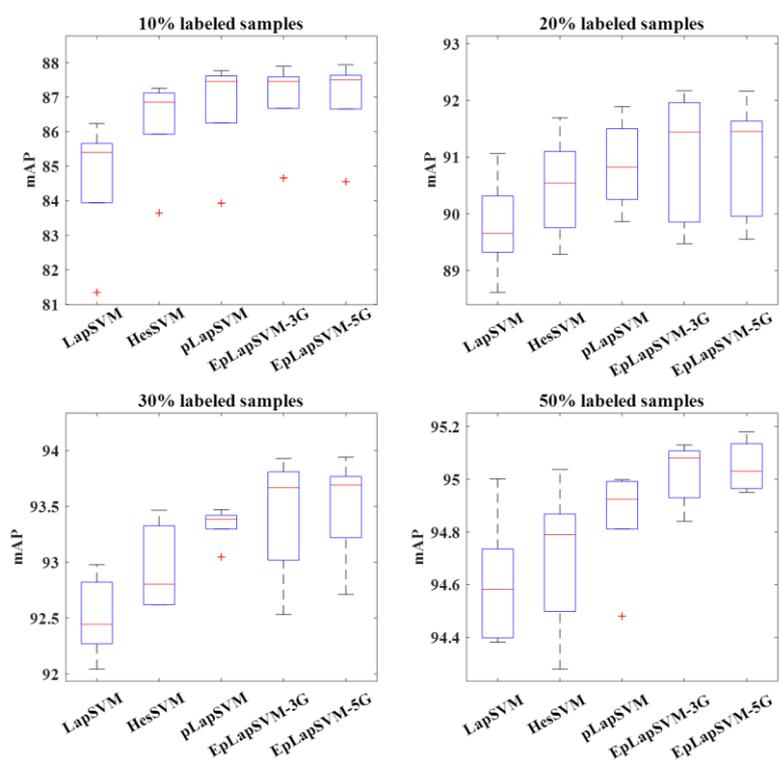

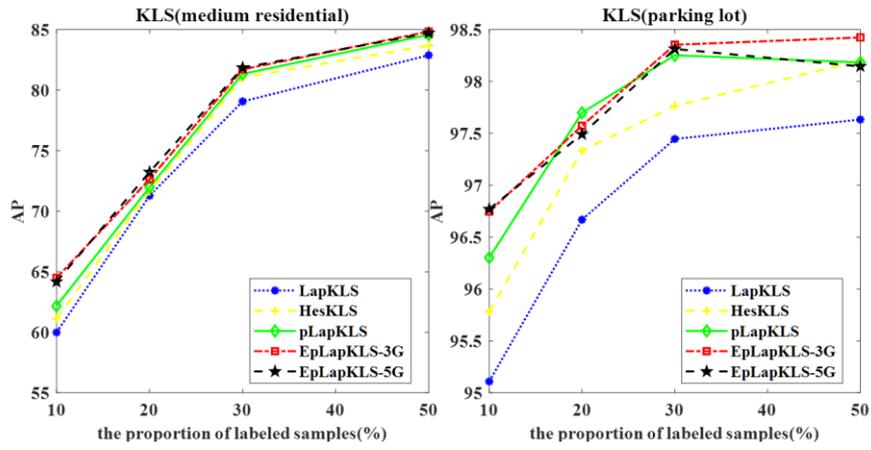

Fig. 5. mAP performance of different algorithms on SVM method.

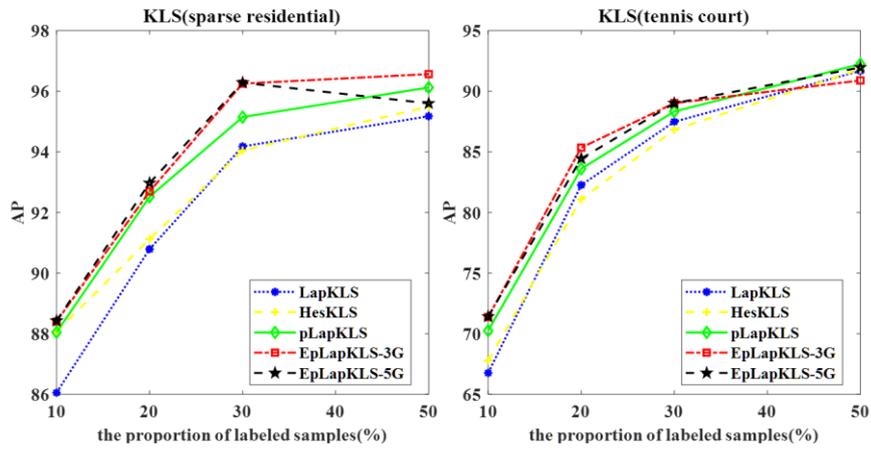

Fig. 6. AP performance of different KLS methods on several classes.

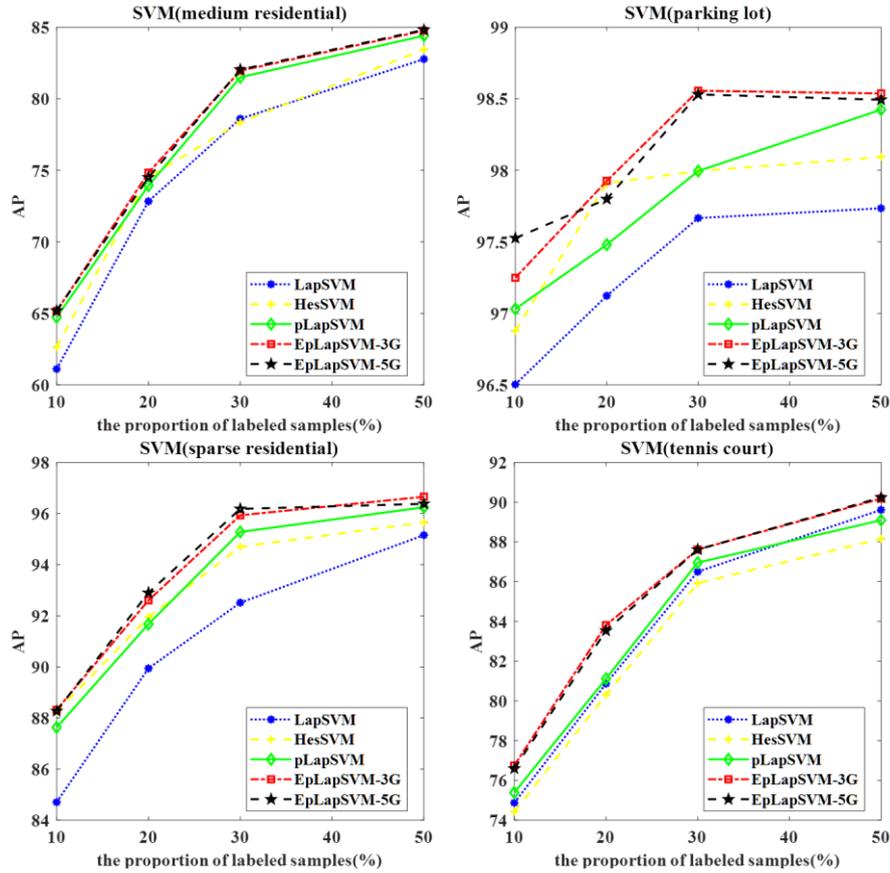

Fig. 7. AP performance of different SVM methods on several classes.

6. Conclusion

As a nonlinear generalization of graph Laplacian, the $p$-Laplacian regularization precisely exploit the geometry of the probability distribution to leverage the learning performance. However, in practical, it is difficult to determine the optimal graph $p$-Lapalcian because the parameter $p$ usually chose by cross validation method which lacks the ability to approximate the optimal solution. Therefore, we propose an ensemble $p$-Laplacian regularization to better approximate the geometry of the data distribution. EpLapR incorporates multiple graphs into a fused graph by an optimization approach to assign suitable weights on different $p$-value graphs. And then, we introduce the optimal fused graph as a regularizer for semi-supervised learning. Finally, we construct the ensemble graph $p$-Laplacian regularized classifiers including EpLapKLS and EpLapSVM for remote sensing image recognition. Experimental results on the UC-Merced dataset show that our proposed EpLapR learner can generalize well than traditional ones.